\pdfoutput=1

\documentclass[11pt]{article}

\usepackage[preprint]{acl}

\usepackage{times}
\usepackage{latexsym}

\usepackage[T1]{fontenc}

\usepackage[utf8]{inputenc}

\usepackage{microtype}

\usepackage{inconsolata}

\usepackage{graphicx}
\graphicspath{ {./images/} }

\usepackage{algorithm}
\usepackage{algpseudocode}

\title{Temporal Alignment of Time Sensitive Facts with Activation Engineering}

\author{
 \textbf{Sanjay Govindan},
 \textbf{Maurice Pagnucco},
 \textbf{Yang Song}
\\
\\
 University of New South Wales, Sydney, Australia \\
 \small{
   \{s.govindan, yang.song1\}@unsw.edu.au, morri@cse.unsw.edu.au
 }
}

\begin{document}
\nolinenumbers
\maketitle
\begin{abstract}
  Large Language Models (LLMs) are trained on diverse and often conflicting knowledge spanning multiple domains and time periods. Some of this knowledge is only valid within specific temporal contexts, such as answering the question, ``Who is the President of the United States in 2022?'' Ensuring LLMs generate time appropriate responses is crucial for maintaining relevance and accuracy. In this work we explore activation engineering as a method for temporally aligning LLMs to improve factual recall without any training or dataset creation. In this research we explore an activation engineering technique to ground three versions of LLaMA 2 to specific points in time and examine the effects of varying injection layers and prompting strategies. Our experiments demonstrate up to a 44\% and 16\% improvement in relative and explicit prompting respectively, achieving comparable performance to the fine-tuning method proposed by \citet{Zhao2024-li}. Notably, our approach achieves similar results to the fine-tuning baseline while being significantly more computationally efficient and requiring no pre-aligned datasets.
\end{abstract}

\section{Introduction}
Large Language Models (LLMs) encode and train on a large corpus of information that the end user can query \citep{Petroni2019-oi,Cohen2024-tb,NIPS2017_3f5ee243}. Their training sets can span a large timeframe leading to overlapping and conflicting answers for time sensitive queries such as “Who is the President of the United States of America?”. Questions like these can have different answers throughout time with the correct answer (Joe Biden in 2024 and Donald Trump in 2025) being temporally sensitive; relevant to the time it is asked \citep{Ge2024-is,Dhingra2022-in,Luu2022-fk}.

Temporally sensitive questions require LLMs to correctly understand the time they are answering for. Without temporal alignment LLMs are recalling facts based on training distributions leading to a chaotic sense of time for factual recall \citep{Zhao2024-li}. This leads to errors such as GPT2-XL recalling that the current Prime Minister of Australia is Malcolm Turnbull. This is at odds with the knowledge the LLM has on hand. The model, if prompted carefully can recall that Scott Morrison is the serving Prime Minister of Australia in 2022, a more recent and temporally relevant answer, and one that supersedes Malcolm Turnbull. The confusion about who the current Prime Minister of Australia is demonstrates how temporal misalignment can lead to erroneous factual recall \citep{Luu2022-fk}.

\begin{figure}
  \centering
  \includegraphics[width=\columnwidth]{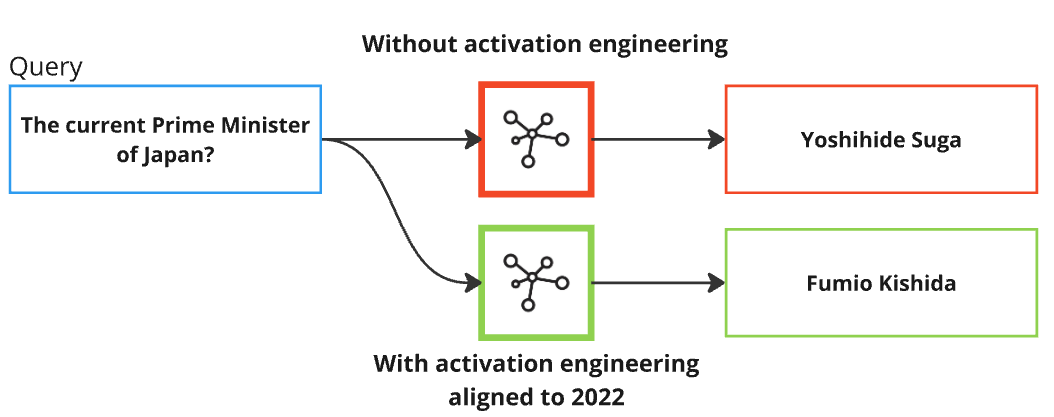}
  \caption{When asked ``Who is the current Prime Minister of Japan'' LLaMA2-7b outputs Yoshihide Suga. Applying activation engineering as temporal alignment assistance for the year 2022 produces the correct set of facts for LLaMA-7b.}
  \label{fig:temporal_alignment}
\end{figure}

 Conflicting answers throughout time can lead to a range of recall errors. To overcome these errors many methods aim to overwrite the subject, relationship, object (SRO) facts held within an LLM \citep{Geva2023-le,Cohen2024-tb,Petroni2019-oi,Yu2023-ll,Dai2022-rx}; for example, overwriting the object associated with the President of the United States of America and altering the probability of the output tokens to preference Donald Trump over Joe Biden. These methods fall under the categories of continual learning \citep{NEURIPS2023_9d8cf124,ke2023continual,jin-etal-2022-lifelong-pretraining}, knowledge editing \citep{de-cao-etal-2021-editing,Yu2024-ig,Hartvigsen2023-wj,Meng2022-ry,Meng2023-rw}, and retrieval augmented generation (RAG) \cite{Lewis2020-wi}, which all strive to provide correct information to the end user.

However, overwriting SRO fact sets ignores the opportunity to realign LLMs temporally to recall the correct fact set that is already known by the model. Building on the work of \citep{Zhao2024-li} we explore the capability of injecting vectors into the residual stream during inference (activation engineering) as a technique to \textit{temporally align models within their existing knowledge cut-off timeframe and correct for relative temporal statements} such as ``Who is the current Prime Minister of Japan?''. Figure \ref{fig:temporal_alignment} demonstrates that when LLaMA2-7b (which has a knowledge cutoff date of September 2022) is asked this question the preferred response is Yoshida Suga (who served as Prime Minister between 2020 and 2021). However, when temporally aligned to the year 2022, using activation engineering the preferred response is Fumio Kishida; who was the current serving Prime Minister of Japan in 2022. The distinction within this scenario is that we are aligning to knowledge the model already possesses. While \citet{Zhao2024-li} uses fine-tuning to align LLMs to a specific year, we explore the effectiveness of activation engineering (AE), which reduces the computational requirements to temporally align, increases the flexibility and responsiveness for end users, and requires less pre-aligned data to reference and train.

In this work we hypothesise that AE is a more efficient method to temporally align models compared to fine-tuning, reducing the amount of training and data required, whilst providing similar outcomes to temporal alignment via fine-tuning. We experiment with the activation methods of \citet{Turner2023-do,Rimsky2024-jv} because of their effectiveness in reducing toxicity, their ease of integration into LLMs and efficiency at inference. While \citet{Turner2023-do} and \citet{Rimsky2024-jv}'s research focuses on qualitative aspects of LLMs such as toxicity and topic fixation, our research on the other hand, aims to understand how AE can affect time-sensitive factual recall, which to the best of our knowledge is a new research perspective. Furthermore, while \citet{Turner2023-do} and \citet{Rimsky2024-jv} look at single layer activation engineering, we explore the capabilities of multi-layer vector injections for aligning time for factual recall.

We apply AE over two datasets, Head of Governments (HOG) and Temporal Alignment Question Answer (Taqa) \citep{Zhao2024-li}. The HOG dataset was created as part of this research to be a small and domain specific dataset. The Taqa dataset \citep{Zhao2024-li} on the other hand was selected to test the effectiveness of AE on a larger, more diverse dataset; exploring the generality of AE on temporal alignment and providing a benchmark to compare against.

We run sweep tests throughout LLaMA2-7b, 13b and 70b models varying the layers of activation and phrases injected into the residual stream. Next, we test the effect of AE on a single layer and then compound the effect by applying AE to multiple layers. We then compare the results of AE against explicit prompting (e.g., In 2022 the President of the United States of America is?), relative prompting (e.g., The current President of the United States of America is?) and fine-tuning; following the methodology of \citet{Zhao2024-li}. We take our findings from the smaller HOG and Taqa-1000 experiments and scale them to the entire Taqa-9000 testing set where we compare our summary Taqa results to \citet{Zhao2024-li} demonstrating a similar alignment to a specific year, but with less computational overhead. 

\section{Related Studies}
\subsection{Temporal alignment}
Temporal alignment \citep{Zhao2024-li} is a relatively new field and distinct from temporal reasoning which aims to understand and influence how time is logically treated by LLMs; such as what date is it from 8 months from now? \citep{tan-etal-2023-towards,Yuan2024-ah}. Temporal alignment on the other hand aims to influence the recall of facts so that they are referenced from a specific point in time providing a time sensitive contextually correct answer.

\citet{Zhao2024-li} is one of the few papers we could find that explores temporal alignment of LLMs. In their research they develop a benchmark dataset Temporal Alignment Question Answer (Taqa) and explored a fine-tuning method for aligning LLMs to a specific year using implicit factual statements. They demonstrate an improvement in alignment and recall of facts from more recent periods compared with explicit and relative prompting. Other methods such as Mend \citep{Mitchell2022-nf}, Memit \citep{Meng2023-rw} and Rome \citep{Meng2022-ry} are focussed on knowledge editing \citep{Yin2024-tc,Ge2024-is,Dong2022-cj,Li2024-hc}, correcting knowledge via hypernetworks or main network editing. These methods overlook the opportunity to correct the fact with minimal intervention via temporal alignment, which can also be used as a tool to minimise the amount of overwriting required in the first place. 

\subsection{Activation Engineering}
We are influenced by the AE approach pioneered by \citet{Rimsky2024-jv} and \citet{Turner2023-do} as their method demonstrates efficiency in steering models towards a desired outcome. Their methods are different to prior AE techniques \citep{Dathathri2019-nb,Hernandez2024-hx} as they use feed forward mechanisms without any pre-training or major alterations to the model. They demonstrate that AE is effective in reducing toxicity and creating topical fixations, producing the desired outcome with minimal intervention.

The AddAct \citep{Turner2023-do} method explores single contrasting pairs to assist in the preference of topic and LLM perspectives. The AddAct method uses simple contrasting statements that do not require any significant data gathering or processing. \citet{Rimsky2024-jv} on the other hand uses a sizable set of crafted contrasting statements which are then added back into the residual stream. We chose to explore \cite{Turner2023-do}'s method due to its efficiency and simplicity in developing activation vectors which are key to quickly altering the temporal alignment of models.  To the best of our knowledge AE has not been explored to correct factual information or align time within LLMs.

\section{Datasets}
\label{sec:datasets}
A variety of datasets have been developed to examine the properties and methods pertaining to LLMs. Recall, biases and reasoning \citep{Thorne2018-ts,Zhong2023-yu} are a few categories of datasets constructed to address research in these areas. These datasets typically look at Subject Relationship Object (SRO) mappings ignore time as a factor for those relationships. For temporal research our dataset requires a record of changes over time for an SRO fact set, expanding the dataset to one that maps SROT: Subject, Relationship, Object and Time. Specifically, we require a dataset that contains consistent time-exclusive answers, where the answer $A$ is only valid between \( t \in [t_s, t_e] \), and $A$ has a consistent set of answers through time.

To the best of our knowledge, there are only two datasets that meet this requirement \citep{Herel2024-pc, Zhao2024-li}. Other datasets such as Atoke \citep{Yin2024-tc}, MQuake \citep{Thorne2018-ts}, ChronoEdit \citep{Ge2024-is} contain question-answer sets through time but do not continuously record the change of answers over time. These datasets are focussed on knowledge editing for time sensitive questions typically exploring one hop knowledge editing effects. For our experiments we have chosen to create a smaller Head of Government (HOG) dataset to provide an easier measure and more efficient experimentation feedback loop. To fully validate our findings are generalisable, we test against Temporal Alignment Question Answer (Taqa) \citep{Zhao2024-li} which contains a much wider knowledge domain. We chose Taqa over \citet{Herel2024-pc} as Taqa has been benchmarked against a fine-tuning alignment approach which we can compare against.

\subsection{Head of Governments (HOG)}
We augment an existing ideologies \citep{Herre2022-xs} dataset to test AE's effect on large periods of time with relatively consistent changes. The domain ``heads of government'' was selected as this information is temporally sensitive and contains common knowledge fact sets such as ``who is the Prime Minister of the UK?'' which can easily be interpreted by LLMs, particularly smaller models such as LLaMA2-7b and 13b. The dataset contains over 175 countries, recording the heads of government between the years 1945 and 2020 inclusive. This set has 175 temporally relevant questions (Who is the current head of government x?), and up to 13,125 explicit temporal questions (In the year Y, who is head of government for X?).

\subsection{Temporal Alignment Question Answer (Taqa) Dataset}
To further validate our temporal alignment experiments we utilise the Taqa dataset \citep{Zhao2024-li}. The Taqa dataset test set contains 9000 temporally relevant questions (e.g., who is the most recent winner of the Stanley Cup?) and up to 113,000 explicit temporal questions (In the year 2010, who was the most recent winner of the nationals skating championship?). The Taqa dataset is purposefully built to exhibit a high rate of answer changes between the years 2020 and 2023, aiming for answers that change at least 5 times over this period.

Whilst Taqa presents a diverse set of question answer pairs, similarities between answers over time and a low recall for LLaMA2 models makes this dataset challenging to work with. It is noted that a range of questions could have very similar answers with just a single token change between the years. These include questions such as, “What edition of the Producers Guild of America Awards was last held?”, whereby the answer is incremented by one in most years. Similarly, we note a series of false positives associated with using F1 token evaluation leading to computational inefficiencies and misleading evaluation. Questions such as “When was the latest Awit Awards ceremony held?”, where the correct answer is a year, such as 2022, leading to any reasonable answer within the 21st century having an F1 score of approximately 0.5. Furthermore, some of these questions could be considered challenging for smaller models such as LLaMA2 7b and 13b which might not have been trained on an extensive body of knowledge.

To improve development efficiency and limit the effect of false positives provided from partial answers we filter for questions that have an F1 Score above 0.5 when answered relatively by LLaMA2-7b. We capture questions LLaMA2-7b, 13b and 70b can answer confidently, creating a more sensitive testing dataset and minimise the testing feedback loop. The Taqa dataset reduces from 9000 to 2930 question answer pairs. We further reduce this to the first 1000 question answer pairs within the 2930 filtered set. This dataset is denoted as Taqa-1000 whilst the full Taqa dataset is denoted as Taqa-9000.

\section{Methods}

\begin{figure*}
  \centering
  \includegraphics[width=425px]{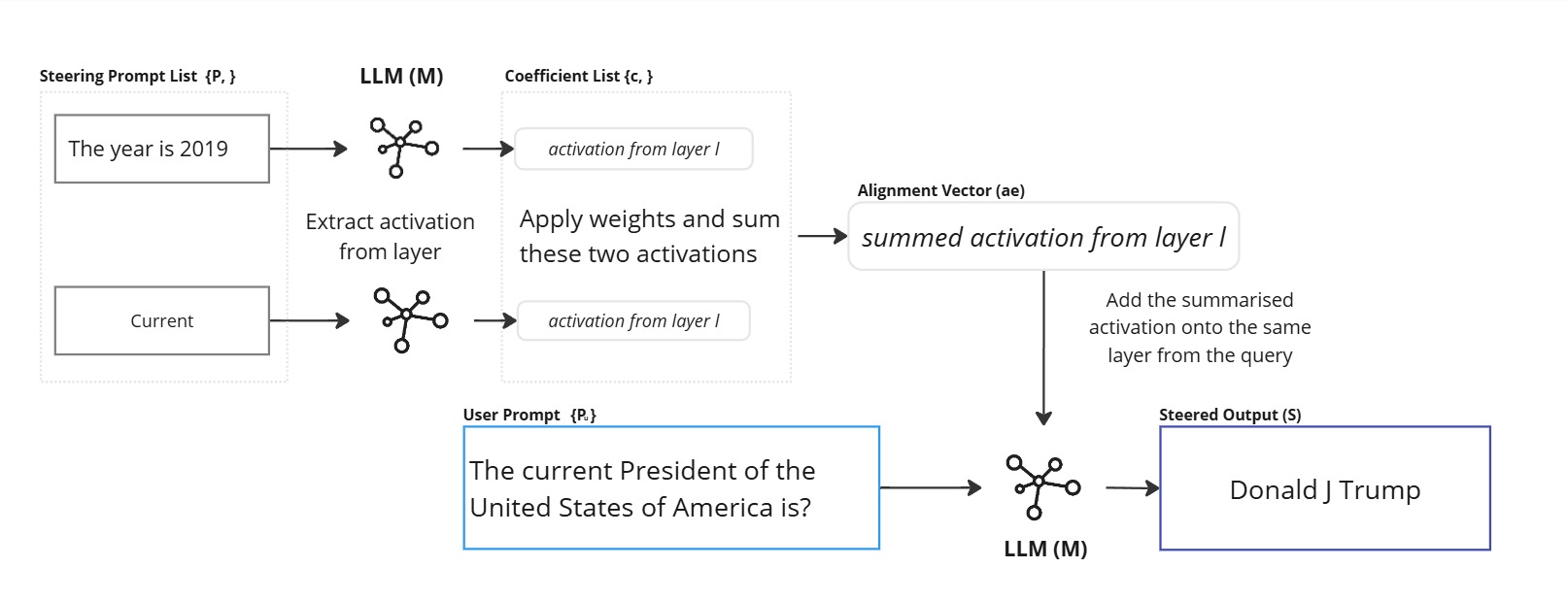}
  \caption{Activation Engineering in LLMs. A set of vectors is extracted from layer \(l\), multiplied by a coefficient and added together. Finally, this vector is added into a temporal question to temporally align the model.}
  \label{fig:activation_engineering}
\end{figure*}
As outlined in Algorithm \ref{alg:activation_engineering} and Figure \ref{fig:activation_engineering}, during the prediction of an answer for a temporally sensitive question \(p_u\) such as, ``Who is the current President of the United States of America?'', we inject a pre-defined steering vector \( ae \) that represents the specific year in to layers \(l\) of the model.

These steering vectors are created from running temporal phrases \(\{p,\}\) such as a year number through the model and extracting the activation vectors \(h\) before a selected layer. Using \(h\), we apply a positive or negative coefficient \(c\) to phrase's vector increasing or reducing the influence of the phrase, and producing \(h_a\). Negative coefficients are used to reduce the influence of a phrase, whilst positive coefficients are used to increase the influence of a phrase upon the output of a model.

Having developed a set of vectors \(h\) that have been multiplied by a coefficient, we sum the set of phrase vectors together to produce a single vector \(ae\) that encapsulates a time period we aim to align to. We add the alignment vector \(ae\) to the residual stream of query prompt \(p_u\) at the same layer \(l\) the original activation vectors \(h\) were extracted from. Ultimately, this small vector ``nudge'' can change the probability of output tokens, steering the model to recall information from a specific period of time. 
 
\begin{algorithm}

\begin{algorithmic}[1]
\Require $\{p,\} =$ steering prompts list
\Require $p_u =$ user prompt
\Require $\{l,\} =$ target layer list
\Require $\{c,\} =$ coefficient list
\Require $a$ = alignment position (front)
\Require $M =$ pre-trained language model
\Ensure $S =$ steered output

\State $mtl = max(len(p)$ for $p$ in $\{p, ...\})$ max token length
\For{each $l$ in $(\{l, ...\})$}
  \State $ae = \{\}$ empty activation vector
  \For{each $p, c$ in $(\{p,\},\{c,\})$}
    \State $(p) \gets$ pad right to match $mtl$
    \State $h \gets M.forward(p).activations[l]$
    \State $h_a \gets h \times c$
    \State $ae \gets ae + h_a$
  \EndFor
  \State $q \gets M.forward(p_u).activations[l]$
  \State $S \gets M.continue\_forward(ae + q@a)$
\EndFor
\end{algorithmic}
\caption{Activation engineering}
\label{alg:activation_engineering}
\end{algorithm}

\textbf{Injection alterations} Our method changes how to inject these vectors into the model, creating a list of layers to inject into \(\{l, ...\}\). We perform two types of injections. Where prior studies \citep{Rimsky2024-jv,Turner2023-do} have focussed on single layer activation engineering we explore the capabilities of multi-layer activation engineering on factual recall. We hypothesise that small nudges throughout a model can provide a more stable outcome for tasks related to factual recall as the vectors are applied to multiple layers providing a more consistent steering of the model. \textbf{Single-layer:} Applying the vector to a single layer only. \textbf{Multi-layer:} Applying vectors additively from layer 4 onwards. e.g., applying vectors to layer 4 and 5, or applying vectors to layer 4, 5 and 6.

\textbf{Temporal prompts} We alter the prompts \($\{p,\}$\) required to create the steering vectors. Specifically we alter the prompt in three ways. \textbf{Year only:} Only the year preferenced for alignment (e.g., 2010). We test year only as a way to efficiently temporally align the model, providing a single year number to see how a basic injection can align the output. A coefficient \(c\) of 4 is applied for single layer and 1 for multi layer. \textbf{Context phrase:} The preference year with context about the number (e.g., the year is 2010). Given the prompting experiments conducted by \citep{Zhao2024-li}, regarding implicit and explicit alignment prompts we broaden our testing to include context about the numbers injected into the residual stream adding ``The year is''. A \(c\) of 4 is applied for single layer and 1 for multi layer. \textbf{Contrasting pair:} A preference year as a positive, and ``recent'' as a negative. e.g., 2021 having a \(c\) of 4, and ``recent'' having a \(c\) of -2 for a single layer and \(c\) of 2 and -1 for multi layer.

\section{Experiments}
\subsection{Models and Prompting}
We use autoregressive decoder only transformers, specifically LLaMA2 7b, 13b and 70b. These models are selected because of their ease of integration with existing tools sets such as AddAct \citep{Turner2023-do} and TransformerLens \citep{nanda2022transformerlens}, as well as their use in \citet{Zhao2024-li}. With regard to setting up the prompts we mimic the setup used by \citet{Zhao2024-li}. For any Taqa question we add a series of non-time sensitive example question answer responses before the Taqa test question. This primes the LLM to reply with a similar short answer when answering the Taqa test question. The question answer examples have no time sensitive facts but instead ask and answer general questions such as ``What is the capital of France?''. When adding these question answer examples for \textit{relative benchmark} tests we keep the question answer examples generic and remove any time from the prompting. For \textit{explicit benchmark} tests we use prompts that reference the year we wish to align to prefixing our question answer examples with ``as of the year x'', where x denotes the year we wish to align to. When testing the effect of AE we use the same question set up scenario as the relative benchmark tests.

\subsection{Evaluation Criteria}
Similar to \citep{Zhao2024-li} and the QA methodologies of \citep{Kwiatkowski2019-gt} we use an averaged F1 and F1 max score for evaluation. An average F1 score demonstrates the effectiveness of aligning on a single year whilst the F1 max score monitors for a significant loss of information over the other years. If the F1 max score decreases drastically compared to benchmark tests, we can assume the method is having an overall negative effect on the model’s output. The F1 max score is calculated between 1945 and 2020 and 2000-2023 for the HOG and Taqa datasets respectively.

\subsection{Baselines}
\begin{figure*}[t]
  \centering
  \includegraphics[width=0.50\linewidth]{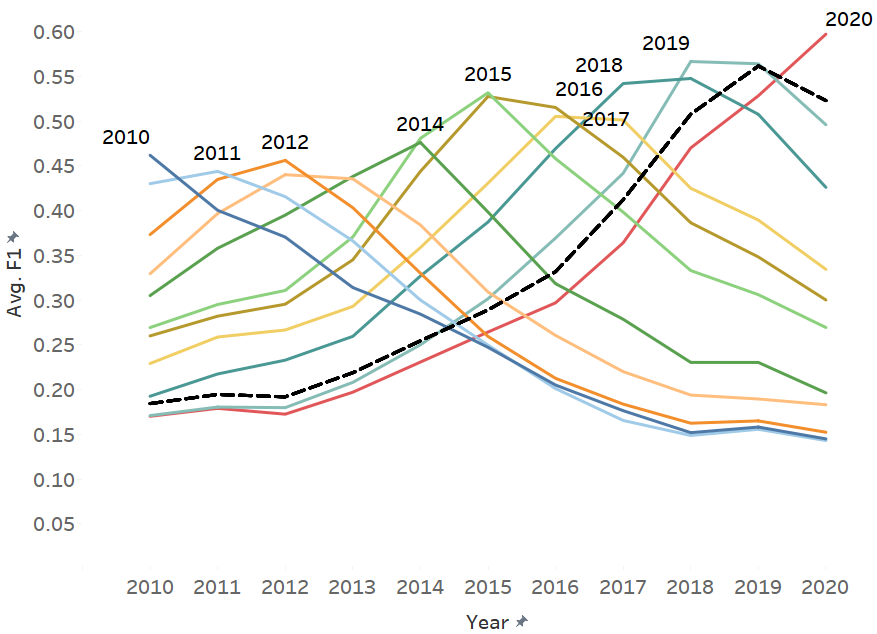}\hfill
  \includegraphics[width=0.50\linewidth]{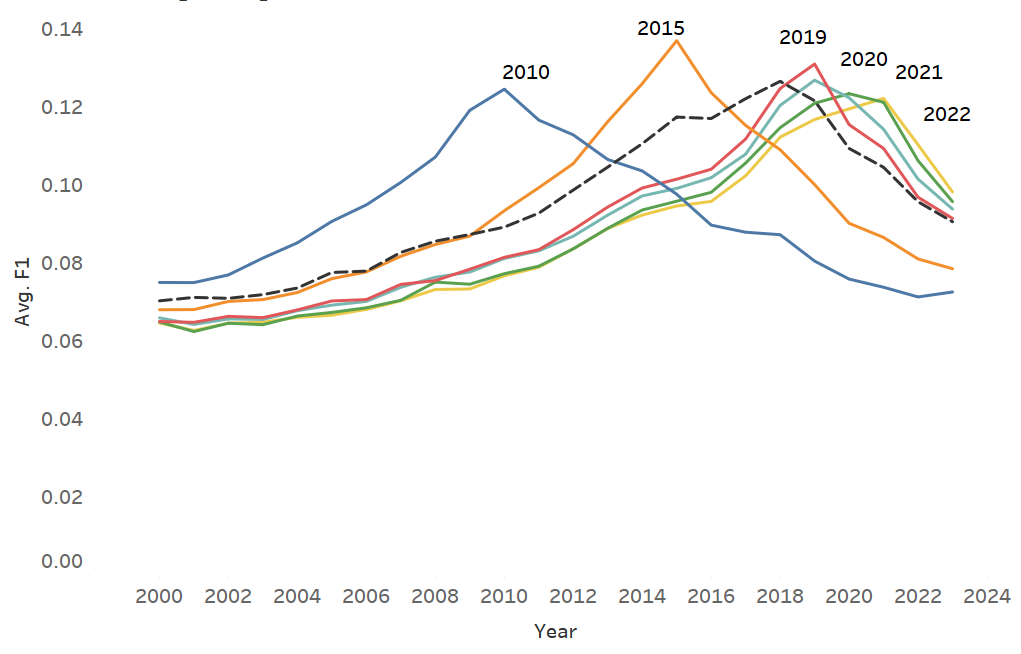}
  \caption{Left (HOG Dataset), right (Taqa-9000) benchmarking F1 scores for LLaMA2-7b for both relative (checked line) and explicit prompts.}
  \label{fig:7b_benchmarking}
\end{figure*}
We first evaluate three baseline approaches, relative prompting, explicit prompting and fine-tuning on the LLaMA models, without involving AE. Relative prompting poise questions such as ``who is the current President of the United States of America?''. Explicit prompting, asks ``In the year 2022, who is the President of the United States of America?'', as an example. Fine-tuning mimics the alignment technique and closely follows the hyperparameters of \citet{Zhao2024-li} and aligns the model to a specific year using implicit factual statements from the Taqa training set. Whilst we aim to mimic the specific techniques of \citet{Zhao2024-li} we were only able to apply full parameter fine-tuning to LLaMA2-7b and 13b models. Due to computational limitations we apply PEFT LoRA \citep{DBLP:journals/corr/abs-2106-09685} fine-tuning to LLaMA2-70b, and reduced the batch size of all training to 8.

Figure \ref{fig:7b_benchmarking}, which focuses on LLaMA2-7b relative and explicit prompting illustrate that the alignment bias of relative and explicit prompts could be related to topics. Earlier years generally produce the worst relative and explicit F1 scores, whilst more recent years exhibit the best F1 scores for the HOG dataset. This isn't the case for the Taqa-9000 dataset which demonstrates 2015 being an easier year to align with explicit prompting. These findings are consistent across 13b and 70b models. The Taqa-9000 dataset illustrates lower performance for explicit prompting alignment for years 2020 to 2022, compared to the gains experienced from explicit prompting in the year 2015. Furthermore, the marginal gains for 13b and 70b compared to 7b between relative and explicit prompting (Table \ref{tab:table_taqa_1000_scores}) most likely stem from the filter conditions applied to Taqa-1000 which favoured question answers pairs that LLaMA2-7b could answer with a high F1 score. On the other hand, using the Taqa-9000 dataset which is a larger and more diverse dataset, it can be seen that in Table \ref{tab:table_taqa_9000_scores} larger models produce better F1 Scores and the gain between explicit and relative scores scales with the size of the model.

\subsection{Results}
\begin{table*}[ht!]
  \centering
  \resizebox{\textwidth}{!}{%
  \begin{tabular}{lccccccc}
    \hline
    & \multicolumn{2}{c}{\textbf{LLaMA 7b}} & \multicolumn{2}{c}{\textbf{LLaMA 13b}} & \multicolumn{2}{c}{\textbf{LLaMA 70b}} \\
    \textbf{}   & \textbf{2021} & \textbf{2022} & \textbf{2021} & \textbf{2022} & \textbf{2021} & \textbf{2022} \\
    \hline
    \textbf{Benchmark}  & & & & \\
    Relative Prompting  & 25.0 & 22.3 & 20.9 & 18.5 & 23.9 & 20.7 \\
    Explicit Prompting  & 26.5 & 23.9 & 23.6 & 23.1 & 29.7 & 27.3 \\
    Fine-tuning & 27.2 & 24.4 & \textbf{28.7} & \textbf{24.5} & 32.9* & 29.4* \\
    \hline
    \textbf{Single layer}  & & & & \\
    Year only  & 28.4 (L6) & 25.9 (L8) & 24.9 (L6) & 23.8 (L10) & 31.0 (L16) & 29.4 (L16) \\
    Context and year & 27.9 (L4) & 26.0 (L6) & 24.5 (L6) & 22.9 (L12 & 30.3 (L4) & 28.2 (L4)) \\
    Contrasting Pair  & 28.6 (L6) & 26.4 (L6)  &  21.6 (L6) & 23.3 (L8) & 31.2 (L16) & 29.7 (L16) \\
    \hline
    \textbf{Multi-layer}   & & & & \\
    Year only & 28.7 (L4-7) & \textbf{28.2 (L4-7)} & 25.0 (L4-10) & 24.2 (L4-10) & 33.8 (L4-20) & 31.1 (L4-20) \\
    Context and year  & 28.6 (L4-8) & 26.0 (L4-9)  & 24.8 (L4-10) & 23.5 (L4-13)  & 33.3 (L4-20) & 31.4 (L4-17) \\
    Contrasting Pair & \textbf{29.0 (L4-10)}  &  26.3 (L4-5)   & \textbf{25.7 (L4-11)} & \textbf{24.4 (L4-11)} & \textbf{34.5 (L4-20)} & \textbf{31.6 (L4-20)} \\
    
    \hline 
    \end{tabular}}
    \caption{F1 scores for single layer and multi-layer experiments for Taqa-1000. ‘L’ denotes which layer(s) the optimal score came from, and * denotes that the fine-tuning method was PEFT LoRA.}
  \label{tab:table_taqa_1000_scores}

\end{table*} 
\begin{table}
  \resizebox{\columnwidth}{!}{%
  \begin{tabular}{lccccccc}
    \hline
    & \multicolumn{2}{c}{\textbf{LLaMA 7b}} & \multicolumn{2}{c}{\textbf{LLaMA 13b}} & \multicolumn{2}{c}{\textbf{LLaMA 70b}} \\
    \textbf{}   & \textbf{2021} & \textbf{2022} & \textbf{2021} & \textbf{2022} & \textbf{2021} & \textbf{2022} \\
    \hline
    \textbf{Benchmark}  & & & & \\
    Relative Prompting  & 10.5 & 9.6 & 10.7 & 9.8 & 14.1 & 13.0 \\
    Explicit Prompting  & 12.2 & 11.0 & 12.2 & 11.7 & 16.7 & 16.0 \\
    Fine-tuning & 12.7 & 12.0 & 14.0 & 13.7 & 19.3* & 17.7* \\
    \hline
    \textbf{Multi-layer}  & & & & \\
    Contrasting Pair  & 13.1 & 12.3 &  13.6 & 12.9 & 20.0 & 19.1 \\    
    \hline 
    \end{tabular}}
    \caption{F1 scores for multi-layer experiments for Taqa-9000. * indicates PEFT LoRA fine-tuning.}
  \label{tab:table_taqa_9000_scores}
\end{table}

\begin{figure*}[hbt!]
  \centering
  \includegraphics[width=0.40\linewidth]{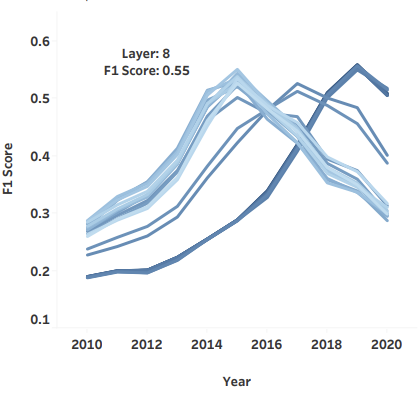}\hfill
  \includegraphics[width=0.40\linewidth]{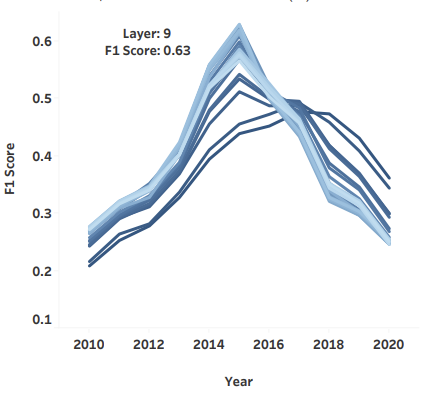}
  \caption{Left (LLaMA2-7b), right (LLaMA2-70b) single layer AE effect on the HOG dataset, using ``year only'' prompting aligning to the year 2015. Layers 4-29 are present. Lighter colours denote lower layers (4-11), and darker colours denote higher layers (12-29). The labels denote the best result and layer.}
  \label{fig:hog_LLaMA_sweep4-29}
\end{figure*}

\begin{figure*}[hbt!]
  \centering 
  \includegraphics[width=0.45\linewidth]{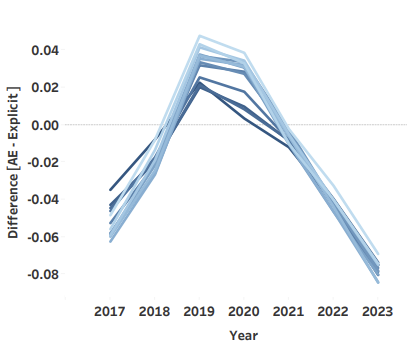}\hfill
  \includegraphics[width=0.45\linewidth]{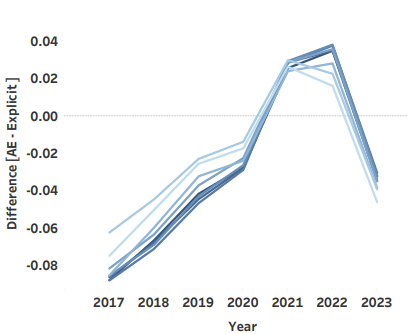}
  \caption{Left (Single layer), right (Multi layer) alignment to 2022 with AE applied to LLaMA2-70b. AE is applied to different layers. The Y-axis is the difference in F1 score between our AE method and explicit prompting. Darker colours denote higher layers. For multi-layer approach, layer 4 is the first layer, and the colour denotes the last layer included. The maximum layer count in both graphs is 26.}
  \label{fig:sensitivity_70b_taqa_1000}
\end{figure*}

As shown in Table \ref{tab:table_taqa_1000_scores} our AE method improves results by up to +10\% points compared to relative prompting and +5\% points compared to explicit prompting. For both the HOG (Appendix \ref{sec:appendix_hog_results}) and Taqa-1000 datasets, we note that multi-layer with a contrasting pair prompt is an effective technique for aligning LLaMA2 13b and 70b models whilst the 7b model benefits from a mixture of alignment techniques. The F1 max scores for all test cases are similar to the relative F1 max scores (Appendix \ref{sec:appendix_hog_results}, \ref{sec:appendix_taqa_1000_results} and \ref{sec:appendix_taqa_9000_results}). This suggests that the loss of information in other time periods is made up by the correction of information from temporal alignment via AE. This is in contrast to the explicit prompting which has a decrease in F1 max score compared to the relative prompt. Overall, Table \ref{tab:table_taqa_9000_scores} demonstrates that AE and fine-tuning produce very similar scores, with differences of up to  $\pm1\%$ between the methods. When examining LLaMA2-70b, we note that AE can outperform the PEFT LoRA fine-tuning method and explicit prompting. We speculate that if PEFT LoRA can achieve a similar score to AE, then full parameter fine-tuning would surpass the best AE F1 scores.

\textbf{Time and computational efficiencies}
Fine-tuning LLaMA2 models (7b, 13b, 70b) for temporal alignment demands significant time and computational resources, whereas AE achieves similar results with lower GPU and time requirements. Generating steering vectors for alignment prompts takes only ~0.05 seconds, enabling AE to quickly align models. In contrast, fine-tuning requires extensive data processing and training time (10–15 minutes for 7b and 13b, ~30 minutes for 70b with PEFT LoRA); for 2 epochs, using \citet{Zhao2024-li} hyperparameters. Fine-tuning also demands substantial GPU power, with configurations ranging from 1xA100-80G (7b), 2xA100-80G (13b) and 3xA100-80G (70b, using PEFT LoRA). AE, however, requires only inference-capable GPUs, with 7b running on a V100-32G, 13b on 1xA100-80G, and 70b on 2xA100-80G. Overall, AE offers a more efficient alternative to fine-tuning for temporal alignment.

\textbf{Layer ablation} To limit the degrees of freedom within our problem space we conducted an ablation study applying AE to individual layers throughout 7b, 13b and 70b parameter models. Specifically we injected the ``year only'' vector, testing individual layers between 4-29 for 7b and 4-39 for 13b and layers 4-29 for 70b. Layers earlier than 4 or later than 29 for 7b and 39 for 13b were deemed inconsequential to test; too low, and the model is still encoding the input, too high and the model is only refining the probability of tokens \citep{Geva2023-le}. For LLaMA2-70b we limited our test to layers 4-29 after noting a consistent drop in performance past layer 26, akin to the performance drops experience at layers 14 onwards for 7b and 13b. We run this sweep test over the HOG dataset and confirm the results when examining the application of AE with the Taqa-1000 dataset.

Figure \ref{fig:hog_LLaMA_sweep4-29} highlights that single layer injection only works for lower layers (4-14 for 7b \& 13b, and 9-24 for 70b), with higher layers (14-29 for 7b \& 13b, 24 and beyond for 70b) seemingly ignoring the activation vector injection. This indicates that the influence of steering vectors for temporal alignment could be capped to first third of LLaMA2 models. The application of AE at higher layers leads to a reversion to the original relative prompt preference year, which is illustrated by Figure \ref{fig:hog_LLaMA_sweep4-29}. Lower layers for all models provide better steering as evidenced in the results of sweep tests from aligned years 2010 to 2020 (Appendix \ref{sec:appendix_different_alignment_years}). The loss of influence exhibited by AE in higher layers is most likely due to a change in layer behaviour \citep{Geva2023-le}; early layers attend to the inputs creating semantic representations. The mid-layers attend to those semantic representations to extract relevant information. Higher layers are tasked with refining the prediction for the next token. For LLaMA2-70b, the application of AE on layers below layer 9 produces results that are worse than AE application to later layers such as 20 and 21. The ineffectiveness of earlier layers is most likely due to the scale of the model, which proportionally has more layers dedicated to encoding user inputs \citep{Geva2023-le}.

\textbf{Multi layer stability} The sweep profiles for single layer AE (Figure \ref{fig:hog_LLaMA_sweep4-29}) demonstrate that sensitivity is only a concern for later layers. From layers 4-12 there is a reduction of -3 from the F1 score of the optimal layer. Beyond layer 12 we note a significant drop in the F1 score for the preferenced year 2015. In contrast, Figure \ref{fig:sensitivity_70b_taqa_1000} highlights how latter layers using a multi layer strategy produce some of the best F1 scores. The multi-layer strategy seemingly reduces the risk of missing the optimal singular layer to influence and suggests that the effect of compounding smaller coefficient additions on the residual stream can increase the overall F1 score.

\section{Conclusion}
Temporal alignment can improve the factual recall accuracy of models but can currently only be achieved through fine-tuning, a computational and time intensive process. We have demonstrated similar results to the fine-tuning process using activation engineering (AE) on LLaMA2-7b, 13b and 70b models. AE can produce similar results to fine-tuning with less data preparation and less upfront computation and time requirements. For smaller models querying question answer pairs the model is historically confident in, AE can provide more accurate outcomes than fine-tuning and explicit prompting, presenting an opportunity for smaller models to integrate AE as a temporal alignment mechanism. Our ablation studies have shown that AE effects a model's perception of time through specific tweaks in the residual stream of lower layers in an LLM. We leave open the idea that AE temporal alignment can be used in conjunction with knowledge hypernetwork methodologies to provide temporal alignment as a mechanism to updated knowledge.

\section{Limitations}
In this study, AE has been applied to isolated LLMs which have no access to external information. The information produced from our tests are a result of the model's pre-training. For LLaMA2-7b, 13b and 70b these models have a knowledge cut off date of September 2022 limiting our AE experiments to this timeframe. This study has not investigated the effect of AE upon external information integration systems into LLMs, such as RAG and knowledge editing with hyper networks.

In addition, we used the hyperparameters defined by \citet{Zhao2024-li} when developing our fine-tuning baseline. We did not optimise these parameters for any difference in our fine-tuning dataset that may arise. Furthermore, this study was limited in its use of larger GPUs to fine tune LLaMA-70b. Our results for LLaMA2-70b fine-tuning was derived from PEFT LoRA fine-tuning which we speculate produced a suboptimal output compared to full parameter fine-tuning.

Finally, we did note that an increase in epochs from 2 to 3 dramatically decreased the number of ``I don't know'' answers. Taqa-1000 dataset, LLaMA2-13b for the year 2022 we note that the relative and explicit benchmarks produced 110 and 64 answers respectively which resulted in ``I don't know''. This is compared to the fine-tuning benchmark which produced 3 and the AE alignment method which produced 94 ``I don't know'' answers. Whilst we note that this change in response might relate to the model becoming overconfident with its answers, we did not fully investigate the extent of this effect across other years and models.

\bibliography{custom}
\clearpage
\onecolumn
\appendix

\begin{center}
\textbf{\LARGE Supplementary material}
\end{center}

\section{Different alignment years}
\label{sec:appendix_different_alignment_years}
\begin{figure*}[hbt!]
  \centering 
  \includegraphics[width=0.5\linewidth]{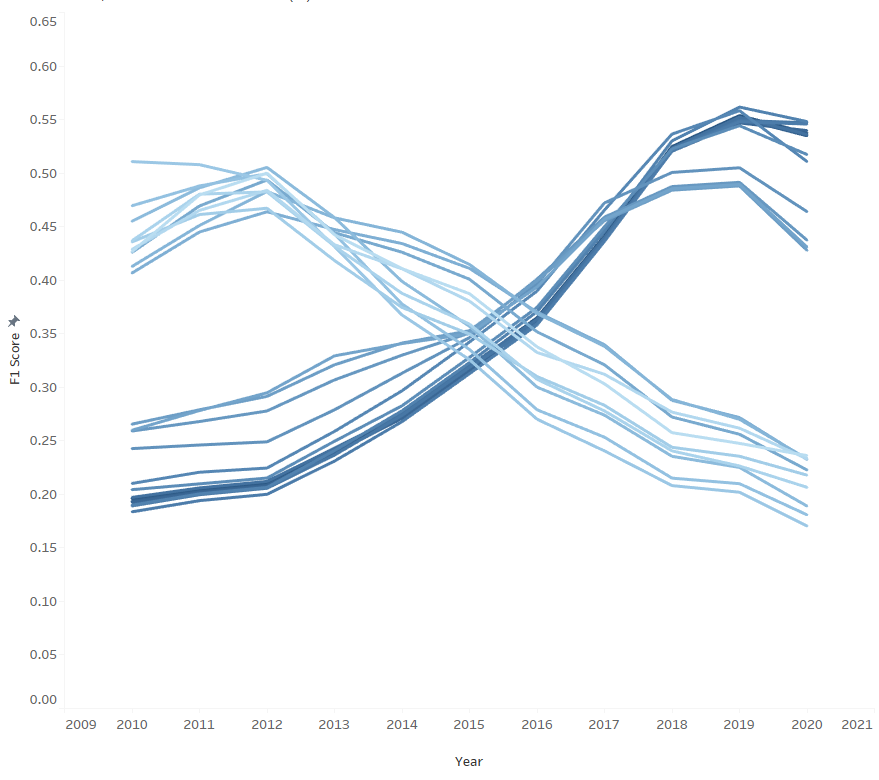}\hfill
  \includegraphics[width=0.5\linewidth]{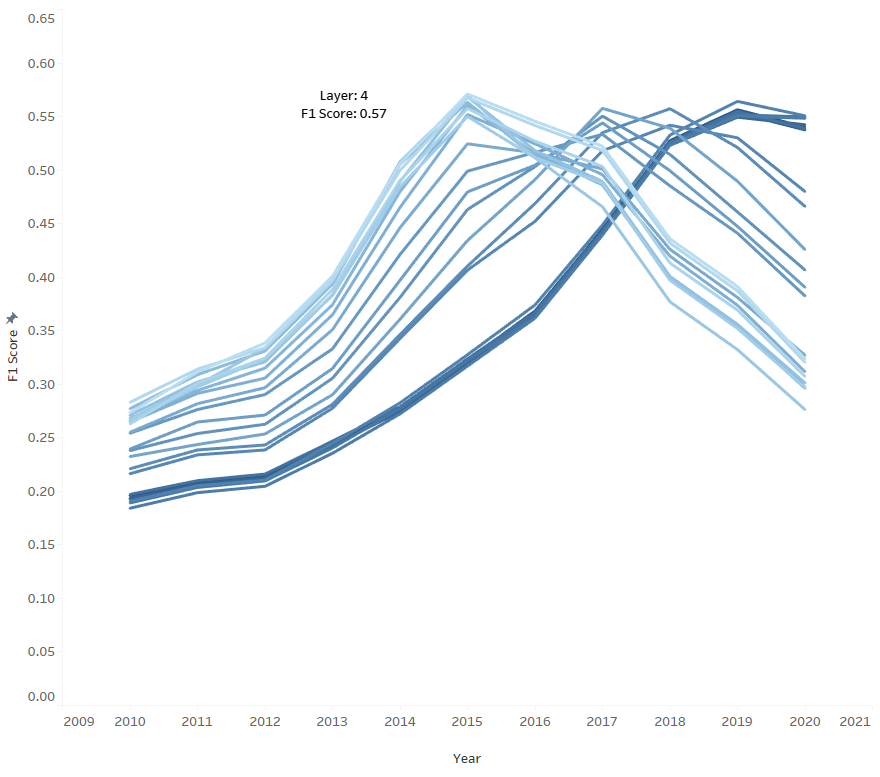}\hfill
  \includegraphics[width=0.5\linewidth]{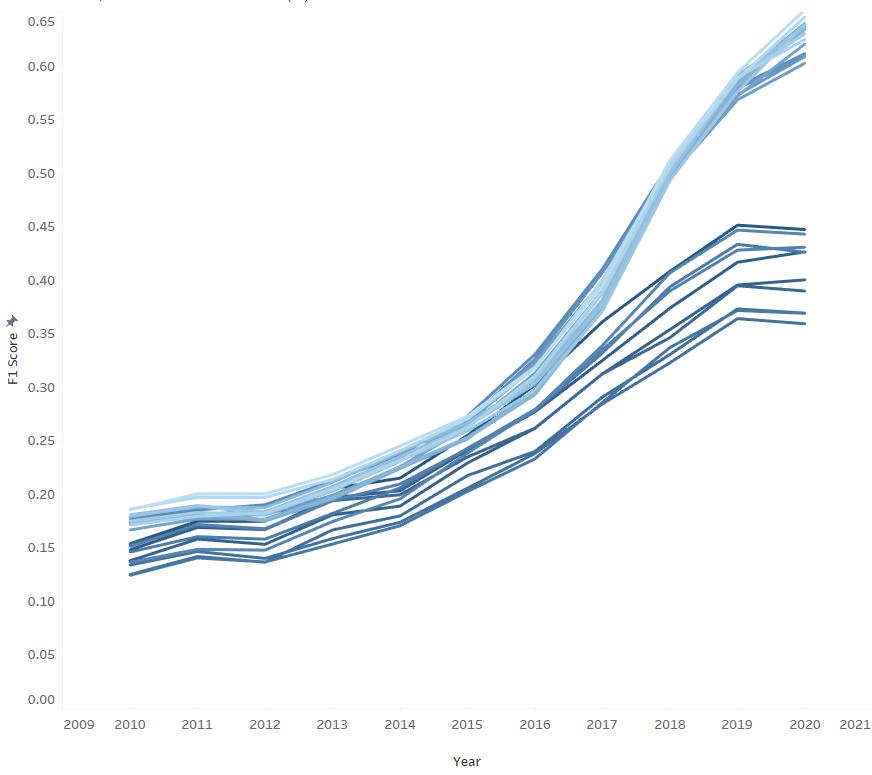}
  \caption{ Hog dataset using LLaMA2-7b aligned to different years, left (2010), middle (2015) and right (2020).Injecting into lower (lighter coloured) layers proves easier to steer the model towards a preference year. Attempting to inject activation vectors into higher layers tends to revrt the model to it's original preference year as defined by the relative prompt.}
  \label{fig:appendix_different_alignment_year_3graphs}
\end{figure*}

\clearpage
\section{Results}
\subsection{Hog Results}
\label{sec:appendix_hog_results}
\begin{table}[hbt!]
  \centering
  \resizebox{\textwidth}{!}{%
  \begin{tabular}{lccccccccc}
    \hline
    & \multicolumn{3}{c}{\textbf{LLaMA 7b}} & \multicolumn{3}{c}{\textbf{LLaMA 13b}} & \multicolumn{3}{c}{\textbf{LLaMA 70b}} \\
    \textbf{} & \textbf{2010}  & \textbf{2015} & \textbf{2020} & \textbf{2010} &\textbf{2015} & \textbf{2020} & \textbf{2010} &\textbf{2015} & \textbf{2020} \\
    \hline
    \textbf{Benchmark} & & & & & & & & & \\
    Relative Prompting  & 18.4 & 28.9 & 52.3 & 18.8 &  32.1 & 55.1 & 20.0 &	33.7 & 43.4 \\
    Explicit Prompting  & 46.2 & 53.2 & 59.7 & 45.3 & 53.2 & 60.4 & 55.6 &	58.9 &	68.3 \\
    \hline
    \textbf{Single layer} & & & & & & & & & \\
    Year only & 50.0 (L8) & 54.4 (L10)  & 58.9 (L4) & 51.1 (L8) & 57.1 (L4) & 65.2 (L4) & 63.0 (L13) & 62.6 (L12) &	65.5 (L8) \\
    Context and year & \textbf{52.6 (L13)} & \textbf{55.7 (L4)} & 57.1 (L4) & 55.9 (L10) & 59.3 (L12) & 64.8 (L10) & 63.0 (L9) &	62.6 (L12) &	64.6 (L13) \\
    Year and recent & \textbf{52.6 (L10)} & 55.1 (L14) & 59.3 (L12) & 55.6 (L10) & 59.3 (L10) & 66.3 (L4) & 61.9 (L10) & 62.6 (L14) & 64.6 (L8) \\
    \hline
    \textbf{Multi-layer}  & & & & & & & & & \\
    Year only & 52.4 (L4-11) & 53.8 (L4-10) & 59.1 (L4-6) & 53.1 L(4-12) & 59.3 (L4-11) & 66.7 (L4-8) 62.6 & (L4-29) & 62.2 (L4-23) & 68.7 (L4-20) \\
    Context and year & 43.1 (L4-5) & 52.7 (L4-8) & 56.1 (L4-6) & 52.9 (L4-10) &  58.1 (L4-9) & 64.8 (L4-5) & 63.4 (L4-20) & 62.4  (L4-20) & 65.5 (L4-29) \\
    year and recent & 49.0 (L4-5) & 54.9 (L4-5) & 60.6 (L4-5) & 56.1 (L4-11) & 61.4 (L4-13) & 67.9 (L4-5) & 63.1 (L4-14) & 63.3 (L4-11) & 69.4 (L4-29) \\
    \hline
  \end{tabular}}
  \caption{Average F1 score for all HOG Results.}
  \label{tab:table_hog_scores}
\end{table} 

\begin{table*}[ht!]
\centering
\begin{tabular}{lccccccccc}
\hline
& \multicolumn{3}{c}{\textbf{LLaMA 7b}} & \multicolumn{3}{c}{\textbf{LLaMA 13b}} & \multicolumn{3}{c}{\textbf{LLaMA 70b}} \\
\textbf{} & \textbf{2010} & \textbf{2015} & \textbf{2020} & \textbf{2010} & \textbf{2015} & \textbf{2020} & \textbf{2010} & \textbf{2015} & \textbf{2020} \\
\hline
\textbf{Benchmark} & & & & & & & & & \\
Relative Prompting & 70.2 & 70.2 & 70.2 & 72.9 & 72.9 & 72.9 & 71.4 & 71.4 & 71.4 \\
Explicit Prompting & 69.5 & 75.4 & 71.2 & 71.4 & 73.6 & 71.3 & 75.5 & 75.2 & 74.8 \\
\hline
\textbf{Single Sweep} & & & & & & & & & \\
Year only & 74.3 & 76.4 & 73.0 & 74.4 & 77.3 & 74.9 & 75.5 & 74.9 & 74.7 \\
Context and year & 73.0 & 76.6 & 72.2 & 74.2 & 74.7 & 74.1 & 77.0 & 75.2 & 73.5 \\
Year - recent & 75.6 & 74.7 & 72.4 & 74.6 & 75.3 & 75.9 & 76.5 & 75.5 & 74.0 \\
\hline
\textbf{Compounding} & & & & & & & & & \\
Year only & 73.5 & 75.4 & 73.2 & 75.9 & 77.0 & 76.0 & 76.8 & 75.9 & 74.8 \\
Context and year & 72.1 & 74.8 & 71.4 & 73.8 & 72.4 & 72.4 & 76.3 & 75.9 & 74.8 \\
Year - recent & 73.7 & 74.7 & 72.1 & 74.6 & 75.6 & 76.0 & 77.5 & 77.8 & 74.4 \\
\hline
\end{tabular}
\caption{F1 max score for HOG dataset.}
\label{tab:appendix_hog_f1maxscore}
\end{table*}
\clearpage

\subsection{Taqa-1000 results}
\label{sec:appendix_taqa_1000_results}
\begin{table}[hbt!]
\resizebox{\textwidth}{!}{%
\begin{tabular}{lccccccccc}
\hline
& \multicolumn{3}{c}{\textbf{LLaMA 7b}} & \multicolumn{3}{c}{\textbf{LLaMA 13b}} & \multicolumn{3}{c}{\textbf{LLaMA 70b}} \\
\textbf{} & \textbf{2020} & \textbf{2021} & \textbf{2022} & \textbf{2020} & \textbf{2021} & \textbf{2022} & \textbf{2020} & \textbf{2021} & \textbf{2022} \\
\hline
\textbf{Benchmark} & & & & & & & & & \\
Relative Prompting & 26.0 & 25.0 & 22.3 & 23.3 & 20.9 & 18.5 & 26.1 & 23.9 & 20.7 \\
Explicit Prompting & 27.2 & 26.5 & 23.9 & 26.0 & 23.6 & 23.1 & 30.2 & 29.7 & 27.3 \\
Fine-Tuning & 28.0 & 27.2 & 24.4 & 28.3 & 27.4 & 24.5 & 29.6* & 32.9* & 29.4* \\
\hline
\textbf{Single layer} & & & & & & & & & \\
Year only & 29.2 (L4) & 28.4 (L6) & 25.9 (L8) & 26.0 (L6) & 24.9 (L6) & 23.8 (L10) & 31.0 (L4) & 31.0 (L16) & 29.4 (L16) \\
Context and year & 29.1 (L6) & 27.9 (L4) & 26.0 (L6) & 25.2 (L4) & 24.5 (L6) & 22.9 (L12) & 30.5 (L8) & 30.3 (L4) & 28.2 (L4) \\
Contrasting Pair & 29.3 (L6) & 28.6 (L6) & 26.5 (L6) & 26.3 (L6) & 25.2 (L6) & 22.3 (L8) & 31.2 (L4) & 31.2 (L16) & 29.7 (L16) \\
\hline
\textbf{Multi layer} & & & & & & & & & \\
Year only & 28.8 (L4-9) & 28.7 (L4-7) & 28.2 (L4-7) & 26.8 (L4-10) & 25.0 (L4-10) & 24.2 (L4-10) & 32.6 (L4-23) & 33.8 (L4-20) & 31.1 (L4-20) \\
Context and year & 28.9 (L4-5) & 28.6 (L4-8) & 26.0 (L4-9) & 25.2 (L4-8) & 24.8 (L4-10) & 23.5 (L4-13) & 32.2 (L4-32) & 33.3 (L4-20) & 31.4 (L4-17) \\
Contrasting Pair & 29.0 (L4-11) & 29.0 (L4-10) & 26.3 (L4-5) & 27.2 (L4-12) & 25.6 (L4-11) & 24.4 (L4-12) & 33.2 (L4-20) & 34.5 (L4-20) & 31.6 (L4-20) \\
\hline
\end{tabular}}
\caption{Average F1 score for Taqa-1000 dataset. The * denotes PEFT LoRA fine-tuning instead of full parameter fine-tuning.}
\label{tab:appendix_taqa_1000_f1score}
\end{table}

\begin{table*}[hbt!]
\centering
\begin{tabular}{lccccccccc}
\hline
& \multicolumn{3}{c}{\textbf{LLaMA 7b}} & \multicolumn{3}{c}{\textbf{LLaMA 13b}} & \multicolumn{3}{c}{\textbf{LLaMA 70b}} \\
\textbf{} & \textbf{2020} & \textbf{2021} & \textbf{2022} & \textbf{2020} & \textbf{2021} & \textbf{2022} & \textbf{2020} & \textbf{2021} & \textbf{2022} \\
\hline
\textbf{Benchmark} & & & & & & & & & \\
Relative Prompting & 83.8 & 83.8 & 83.8 & 59.8 & 59.8 & 59.8 & 65.5 & 65.5 & 65.5 \\
Explicit Prompting & 70.1 & 69.2 & 68.3 & 56.9 & 58.2 & 57.9 & 59.3 & 57.7 & 55.1 \\
Fine-Tuning & 76.8 & 74.5 & 75.4 & 64.6 & 65.0 & 64.1 & 71.1* & 70.7* & 70.2* \\
\hline
\textbf{Single Sweep} & & & & & & & & & \\
Year Only & 76.6 & 74.5 & 72.9 & 57.6 & 58.4 & 56.9 & 64.3 & 68.4 & 62.9 \\
Context and Year & 76.9 & 76.6 & 75.5 & 58.3 & 58.7 & 59.2 & 65.9 & 68.0 & 67.7 \\
Contrasting Pair & 76.3 & 74.4 & 74.2 & 58.5 & 59.2 & 57.0 & 68.0 & 64.7 & 62.7 \\
\hline
\textbf{Multi layer} & & & & & & & & & \\
Year Only & 74.5 & 75.0 & 73.4 & 58.2 & 58.3 & 58.1 & 67.3 & 66.3 & 65.0 \\
Context and Year & 78.7 & 74.1 & 72.8 & 57.4 & 57.5 & 57.4 & 67.0 & 66.4 & 65.7 \\
Contrasting Pair & 75.1 & 74.1 & 74.7 & 60.1 & 59.3 & 58.1 & 67.5 & 65.9 & 64.9 \\
\hline
\end{tabular}
\caption{F1 max score for Taqa-1000 dataset. The * denotes PEFT LoRA fine-tuning.}
\label{tab:appendix_taqa_1000_f1maxscore}
\end{table*}
\clearpage

\subsection{Taqa-9000 results}
\label{sec:appendix_taqa_9000_results}
\begin{table*}[hbt!]
\centering
\begin{tabular}{lccccccccc}
\hline
& \multicolumn{3}{c}{\textbf{LLaMA 7b}} & \multicolumn{3}{c}{\textbf{LLaMA 13b}} & \multicolumn{3}{c}{\textbf{LLaMA 70b}} \\
\textbf{} & \textbf{2020} & \textbf{2021} & \textbf{2022} & \textbf{2020} & \textbf{2021} & \textbf{2022} & \textbf{2020} & \textbf{2021} & \textbf{2022} \\
\hline
\textbf{Benchmark} & & & & & & & & & \\
Relative Prompting & 10.9 & 10.5 & 9.6 & 11.6 & 10.7 & 9.8 & 14.9 & 14.1 & 13.0 \\
Explicit Prompting & 12.2 & 12.2 & 11.0 & 12.8 & 12.2 & 11.7 & 17.4 & 16.7 & 16.0 \\
Fine-tuning & 13.6 & 12.7 & 12.0 & 14.4 & 14.0 & 13.7 & 17.7* & 19.3* & 17.7* \\
\hline
\textbf{Multi layer} & & & & & & & & & \\
Contrasting Pair & 13.4 & 13.1 & 12.3 & 13.9 & 13.6 & 12.9 & 19.8 & 20.0 & 19.1 \\
\hline
\end{tabular}
\caption{Average F1 score for Taqa-9000 dataset. The * denotes PEFT LoRA fine-tuning.}
\label{tab:appendix_taqa_9000_f1score}
\end{table*}

\begin{table*}[hbt!]
\centering
\begin{tabular}{lccccccccc}
\hline
& \multicolumn{3}{c}{\textbf{LLaMA 7b}} & \multicolumn{3}{c}{\textbf{LLaMA 13b}} & \multicolumn{3}{c}{\textbf{LLaMA 70b}} \\
\textbf{} & \textbf{2020} & \textbf{2021} & \textbf{2022} & \textbf{2020} & \textbf{2021} & \textbf{2022} & \textbf{2020} & \textbf{2021} & \textbf{2022} \\
\hline
\textbf{Benchmark} & & & & & & & & & \\
Relative Prompting & 38.0 & 38.0 & 38.0 & 34.7  & 34.7 & 34.7 & 43.0 & 43.0 & 43.0 \\
Explicit Prompting & 35.3 & 34.6 & 34.6 & 33.7 & 34.0 & 34.0 & 37.7 & 37.0 & 35.5 \\
Fine-tuning & 40.9 & 41.1 & 41.0 & 39.2 & 40.0 & 39.0 & 47.0* & 47.1* & 46.8* \\
\hline
\textbf{Multi layer} & & & & & & & & & \\
Contrasting Pair & 38.4 & 38.0 & 37.9 & 36 & 35.9 & 31.5 & 44.3 & 43.9 & 42.4 \\
\hline
\end{tabular}
\caption{F1 max score for Taqa-9000 dataset. The * denotes PEFT LoRA fine-tuning.}
\label{tab:appendix_taqa_9000_f1maxscore}
\end{table*}

\clearpage

\section{Efficiency}
\label{sec:efficiency}
Investigating efficiencies of AE over fine-tuning we assessed the wall time required for a set of inferences. We compare relative and AE alignment methods over the use of single layer, and multi layer sweeping to assess the wall time difference between these techniques. Overall we note that a single layer injection increases the inference time by 1.2x whilst multi layer injection increases the time by ~2.5x. The fine-tuning process requires a large amount of data to be processed and a large amount of time spent searching for hyperparameters and fine-tuning the model; our experience indicated at least 5–20 minutes for smaller models such as LLaMA 7b and 13b and up to 30 minutes for LLaMA2-70b training with PEFT LoRA on 3 H100 94G GPUs.

For 7b and 13b models we conduct our testing on a single  A100 80G card. For the 70b model, we conduct our testing on 2 A100 80G cards. We note that the development of a single vector made of a contrasting ``year and recent'' statement is a once off overhead which can be computed on average in under 0.05 seconds. Figure \ref{tab:inference_time} demonstrates the wall time required for the application of this vector into 7b, 13b and 70b models across a single layer and multi-layer application using the 'contrasting prompt'.
\begin{table}[hbt!]
  \centering
  \begin{tabular}{lccc}
    \hline
    \textbf{Model} & \textbf{Relative Prompt (s)} & \textbf{Single Layer (s)} & \textbf{Multi Layer (s)} \\
    \hline
    LLaMA2-7b & 0.69 & 0.91 & 1.86 \\
    LLaMA2-13b & 0.87 & 1.05 & 2.35 \\
    LLaMA2-70b & 1.95 & 2.29 & 4.91 \\
    \hline
  \end{tabular}
  \caption{Wall time (seconds) required for inference with different activation engineering methods for a statement with 10 tokens.}
  \label{tab:inference_time}
\end{table}

\end{document}